\documentclass{article}

\usepackage[nonatbib,preprint]{neurips_2025}

\usepackage[utf8]{inputenc} 
\usepackage[T1]{fontenc}    
\usepackage{hyperref}       
\usepackage{url}            
\usepackage{booktabs}       
\usepackage{amsfonts}       
\usepackage{nicefrac}       
\usepackage{microtype}      
\usepackage{xcolor}         

\usepackage{graphicx}
\usepackage[numbers]{natbib}

\usepackage{amsmath}
\usepackage{amssymb}
\usepackage{mathtools}
\usepackage{amsthm}

\usepackage[capitalize,noabbrev]{cleveref}

\usepackage{algorithm}
\usepackage{algorithmic}

\usepackage{caption}
\usepackage{subcaption}

\usepackage{booktabs}
\usepackage{braket}
\usepackage{bbm}
\usepackage{float}

\usepackage{wrapfig}

\usepackage[framemethod=TikZ]{mdframed}
\usepackage{lipsum} 

\newmdenv[
  linecolor=black,
  linewidth=1pt,
  roundcorner=5pt,
  backgroundcolor=gray!10,
  nobreak=false,
  skipabove=12pt,
  skipbelow=12pt,
]{myframe}

\theoremstyle{plain}
\newtheorem{theorem}{Theorem}[section]

\theoremstyle{definition}
\newtheorem{definition}[theorem]{Definition}

\theoremstyle{remark}

\title{DICE: Dynamic In-Context Example Selection in LLM Agents via Efficient Knowledge Transfer}

\author{%
  Ruoyu Wang \thanks{Corresponding author} \\
  UNSW Sydney\\
  \texttt{ruoyu.wang5@unsw.edu.au} \\
  \And
  Junda Wu \\
  UC San Diego \\
  \texttt{juw069@ucsd.edu} \\
  \AND
  Yu Xia \\
  UC San Diego \\
  \texttt{yux078@ucsd.edu} \\
  \And
  Tong Yu \\
  Adobe Research \\
  \texttt{tyu@@adobe.com} \\
  \And
  Ryan A. Rossi \\
  Adobe Research \\
  \texttt{ryrossi@adobe.com} \\
  \And
  Julian McAuley \\
  UC San Diego \\
  \texttt{jmcauley@ucsd.edu} \\
  \And
  Lina Yao \\
  UNSW Sydney, CSIRO's Data 61 \\
  \texttt{lina.yao@data61.csiro.au} \\
}

\begin{document}

\maketitle

\begin{abstract}
Large language model based agents, empowered by in-context learning (ICL), have demonstrated strong capabilities in complex reasoning and tool-use tasks. However, existing works have shown that the effectiveness of ICL is highly sensitive to the choice of demonstrations, with suboptimal examples often leading to unstable or degraded performance. While prior work has explored example selection, including in some agentic or multi-step settings, existing approaches typically rely on heuristics or task-specific designs and lack a general, theoretically grounded criterion for what constitutes an effective demonstration across reasoning steps. Therefore, it is non-trivial to develop a principled, general-purpose method for selecting demonstrations that consistently benefit agent performance. In this paper, we address this challenge with \textsc{DICE}, \textsc{D}ynamic \textsc{I}n-\textsc{C}ontext \textsc{E}xample Selection for LLM Agents, a theoretically grounded ICL framework for agentic tasks that selects the most relevant demonstrations at each step of reasoning. Our approach decomposes demonstration knowledge into transferable and non-transferable components through a causal lens, showing how the latter can introduce spurious dependencies that impair generalization. We further propose a stepwise selection criterion with a formal guarantee of improved agent performance. Importantly, DICE is a general, framework-agnostic solution that can be integrated as a plug-in module into existing agentic frameworks without any additional training cost. Extensive experiments across diverse domains demonstrate our method’s effectiveness and generality, highlighting the importance of principled, context-aware demo selection for robust and efficient LLM agents.
\end{abstract}

\section{Introduction}

Large language model based agents, powered by techniques such as chain-of-thought prompting \cite{wei2022chain} and agentic frameworks, have become popular and effective for complex reasoning and tool-use tasks \cite{wang2024survey}. A key component of these frameworks is in-context learning (ICL), where a small set of demonstrations is prepended to the prompt so the model can infer the desired behaviour without any parameter updates \cite{brown2020language}. In agentic frameworks such as ReAct \cite{yao2023react}, demonstrations serve to illustrate how the agent should reason and act based on observations, invoke external tools, and interpret results before proceeding to the next step. By showcasing sequences of thought, action, and observation, ICL guides the agent’s policy and enables effective reasoning in novel environments.

While In-Context Learning (ICL) often achieves strong performance, prior work has shown that its effectiveness is highly sensitive to the choice of examples \cite{zhang2022active}. To address this instability, some methods have been proposed for active example selection in ICL, aiming to identify more informative or representative exemplars. Some of these approaches focus on one-step question answering tasks \cite{zhang2022active, margatina2023active}, while some recent work extended it to multi-step, agentic settings \cite{lutz2024wilbur, yang2024cops}. However, these methods typically rely on heuristics or task-specific strategies, and lack a general, theoretically grounded criterion for what constitutes an effective demonstration across reasoning steps of an agent. This limits their generalizability, particularly in specialized or novel environments where suitable demonstrations may be scarce or hard to identify.

\begin{figure*}
  \centering
  \includegraphics[width=1\linewidth]{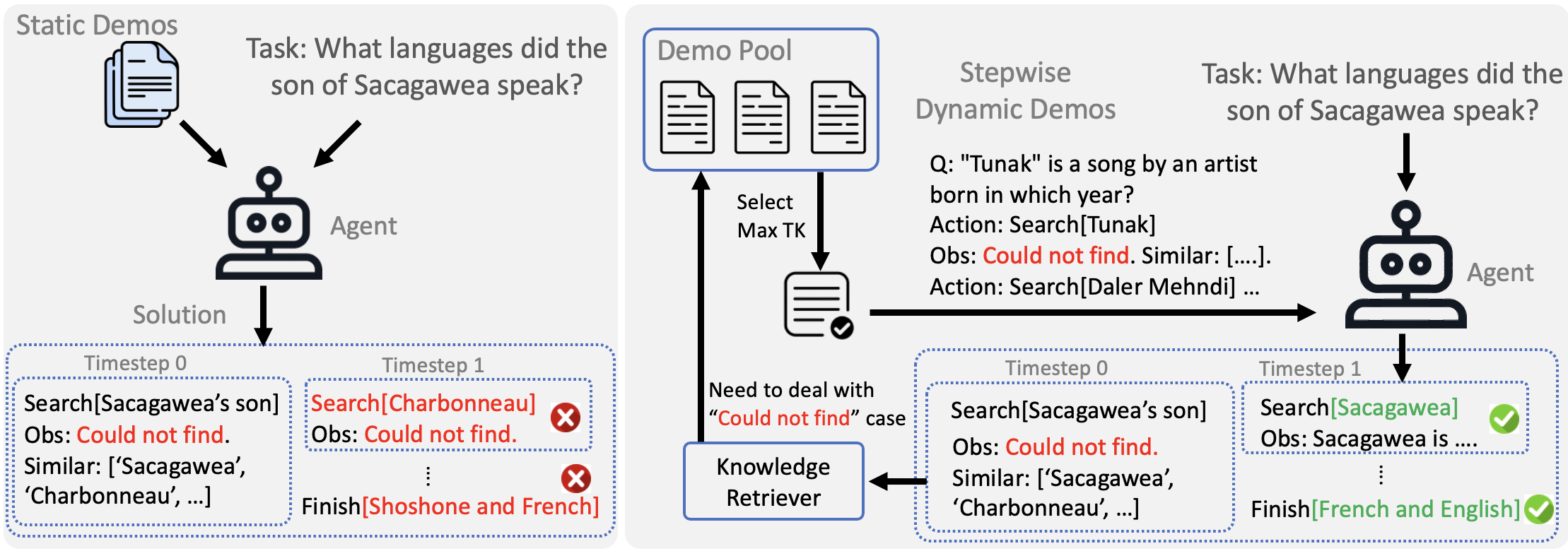}
  \caption{Overview of our stepwise dynamic demonstration selection framework. \textbf{Left}: Without relevant demonstrations, the agent struggles when encountering unfamiliar cases—e.g., failing to deal with "Could not find" case - leading to incorrect actions and ultimately a wrong answer. \textbf{Right}: With DICE, the agent retrieves contextually relevant examples at each time step by maximizing transferable knowledge (TK). When encountering the "Could not find" case, a relevant demo is selected and guides the agent on handling it, enabling successful completion.}
  \label{fig:workflow}
  \vspace{-10pt}
\end{figure*}

Therefore, it is non-trivial to investigate what constitutes a good demonstration for an LLM-based agent, what knowledge should or should not be transferred from demonstrations to the query task, and to develop a principled, general-purpose method for efficient example selection and knowledge transfer that consistently enhances agent performance across reasoning steps. To address these limitations in the existing literature, we propose DICE, Dynamic In-Context Example Selection, a theoretically grounded ICL framework tailored for agentic tasks. DICE enables agents to dynamically select the most relevant demonstrations at each time step of the problem-solving process, by quantifying the amount of transferable knowledge each demonstration provides  (Figure~\ref{fig:workflow}). This allows the agent to overcome the limitations of static example selection and to mitigate spurious correlations introduced by irrelevant or misleading demonstrations. In contrast to prior approaches that require additional model training to guide demonstration selection \cite{zhang2022active,rubin2021learning,ye2023compositional}, DICE is entirely training-free. It is a general, framework-agnostic solution that can be seamlessly integrated as a plug-in module into existing agentic frameworks without incurring any additional training cost.

Specifically, we begin by formalizing the influence of in-context demonstrations on an agent’s decision-making process through a causal perspective. We decompose the knowledge conveyed by demonstrations into transferable and non-transferable components, demonstrating that the latter can introduce spurious dependencies that hinder generalization. Building on this insight, we propose a dynamic demonstration selection algorithm that, at each step of the problem-solving process, identifies the most relevant examples—those that maximize the transferable knowledge beneficial to the current reasoning step. Crucially, our method is supported by theoretical guarantees, establishing improved bounds on the generalization gap. Our main contributions are as follows:
\begin{itemize}
  \item We articulate a causal perspective on the empirical instability of ICL, highlighting how demonstrations can degrade performance by introducing spurious association into the decision-making process. Leveraging this insight, we propose a demonstration selection criterion that mitigates such spurious dependencies by dynamically adapting exemplars to maximize the transferable knowledge relevant to the agent’s subtask at each reasoning step.
  \item We develop a practical strategy to implement our demonstration selection criterion and instantiate it as a plug-in module compatible with existing agentic frameworks. Our method operates entirely at inference time, requires no additional training, and consistently achieves strong empirical performance across diverse domains and agent architectures.
  \item We theoretically show that our selection strategy yields tighter generalization bounds by identifying transferable knowledge, leading to improved performance guarantees.
\end{itemize}

\section{Method}

\subsection{When Demonstrations Impede Agent Performance?} 
\label{method_motivation}

\begin{wrapfigure}{R}{0.3\textwidth}
    \vspace{-10pt}
    \centering
    \includegraphics[width=0.3\textwidth]{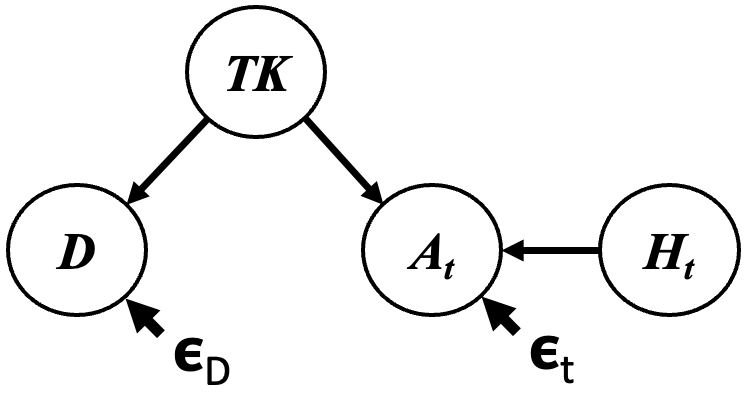}
    \caption{Causal graph showing how demo-specific noise $\epsilon_D$ affects the next action $A_t$ in ICL, introducing spurious correlations that can hinder agent performance.}
    \label{fig:causal_graph}
    \vspace{-10pt}
\end{wrapfigure}

Intuitively, when using in-context learning (ICL) to guide an agent, the provided examples inevitably contain a mixture of knowledge—some of which is relevant and transferable to the new task, and some of which is not. While the transferable knowledge can be transferred to the query task and help the agent perform better, the irrelevant or task-specific information embedded in the demonstrations may introduce spurious cues. These cues can mislead the agent and ultimately hinder performance on the target task. 

We illustrate this intuition using a causal graph, shown in Figure~\ref{fig:causal_graph}, where \( H_t \) represents the agent’s \emph{history} up to the current timestep, \( A_t \) denotes the agent’s next action, \( D \) is the demonstration provided to the agent, and \( \mathit{TK} \) (Transferable Knowledge) refers to the subset of information within \( D \) that is genuinely relevant and beneficial for solving the current decision at step \( t \). In contrast, \( \epsilon_D \) and \( \epsilon_t \) represent task-specific or spurious knowledge that is irrelevant to other tasks and should ideally not influence decisions beyond their original context. Since \( \mathit{TK} \) captures knowledge shared between the demonstration and the current task, it serves as a common cause of both \( D \) and \( A_t \), forming the structure \( D \leftarrow \mathit{TK} \rightarrow A_t \). Additionally, \( H_t \rightarrow A_t \) holds because the agent’s decision at step \( t \) should logically depend on its accumulated experience up to that point. Finally, the task-specific noise terms \( \epsilon_D \) and \( \epsilon_t \) influence only their respective tasks and should not generalize.

This graph intuitively explains a limitation of standard In-Context Learning (ICL). By providing the demonstration \( D \) as input, we inherently condition on \( D \), which opens a \emph{collider} structure: \( \epsilon_D \rightarrow D \leftarrow \mathit{TK} \rightarrow A_t \). As a result, a backdoor path is created, allowing spurious task-specific information from \( \epsilon_D \) to influence the generation of \( A_t \). This unintended information flow can lead the agent to rely on irrelevant cues, ultimately limiting its performance on the target task.

\subsection{DICE: Dynamic In-Context Example Selection}
\label{method_formulation}

To mitigate the limitation described above, we propose a dynamic demonstration selection mechanism that controls the influence of irrelevant knowledge during in-context learning. The core idea is to adaptively select demonstrations most relevant to the agent’s current decision point, thereby reducing the impact of spurious correlations introduced by task-specific information in unrelated examples.

In a standard LLM‑based agent framework, an agent interacts with an environment over discrete time steps $t$. A $Task$ (e.g., a natural‑language instruction) and a set of in‑context demonstrations $\mathrm{Demos}$ are provided. At each step \( t \), the agent takes an action \( a_t \in \mathcal{A} \) and receives an observation \( o_t \in \mathcal{O} \) from the environment, and updates its context \( H_t = ( \mathrm{Demos}, \mathrm{Task}, a_1, o_1, \dots, a_t, o_t) \) until the agent emits a special \emph{Finish} action or a time limit $T$ is reached.

In contrast, our framework render the demonstration set \(\mathrm{Demos}\) dynamic. Specifically, let \(\mathcal{D} = \{ d_i \}_{i=1}^N\) with \( d_i  = \{\text{Task}_i,  a_{i,1},  o_{i,1}, \dots, a_{i,T},  o_{i,T}\}\) be a pool of \(N\) complete trajectories, each illustrating a correct solution to its respective task. At each time step \(t\), in addition to appending the new action–observation pair \((a_t, o_t)\) to the context \(H_t\), we replace the previous \(\mathrm{Demos}\) by selecting a fresh subset of \(M\) trajectories \( \{ d_j \}_{j=1}^M \in \mathcal{D}\) via a retrieval policy \(\pi\), which aims to maximize the expected cumulative task reward:
\[
\pi^* = \arg\max_{\pi} \mathbb{E}\left[ \sum_{t=1}^T \mathcal{R}(a_t) \right],
\]
where \(\mathcal{R}(a_t)\) measures task‑specific performance. By tailoring demonstrations to the agent’s evolving context, our approach filters out irrelevant examples, emphasises transferable knowledge \( \mathit{TK} \), and boost adaptability while mitigating spurious correlations at each decision step.

To enable such a dynamic ICL framework, an automatic demonstration selection criterion is needed. Therefore, we propose a criterion grounded in the Information Bottleneck (IB) principle, which formalizes the trade-off between preserving useful information and discarding irrelevant signals. We define the objective as follows:
\begin{equation}
\label{eq:criterion}
\arg\min_{d_i \in D} J_i, \quad \text{where } J_i = I(d_i; \text{TK}_{d_i}) - \beta I(\text{TK}_{d_i}; A_t)
\end{equation}
where the first term penalizes the inclusion of excessive task-specific details that do not generalize across tasks, and the second term, scaled by a regularization coefficient \( \beta > 0 \), encourages the selection of demonstrations that maximize the mutual information between the agent’s next action \( A_t \) and the transferable knowledge \( \mathit{TK} \).

This objective naturally arises from our causal analysis in Section~\ref{method_motivation}: by maximizing the influence of the causal path \( \mathit{TK} \rightarrow A_t \) while minimizing the effect of the spurious dependency induced by the collider \( \epsilon_D \rightarrow D \leftarrow \mathit{TK} \rightarrow A_t \), our method selectively filters out non-transferable knowledge. In doing so, it enables more effective and generalizable in-context reasoning, particularly in multi-step tasks where compounding errors from irrelevant information can significantly degrade performance.

\subsection{Theoretical Guarantee}
\label{method_theory}
We conduct a theoretical analysis for the method proposed in Section~\ref{method_formulation}. Specifically, we address two key questions: (1) Why is compressing demonstrations to transferable knowledge (\(\mathit{TK}\)) beneficial? and (2) Why does selecting demonstrations using our proposed criterion improve agent performance? To address these questions, we begin by introducing the notion of the \emph{generalization gap}, a metric that quantifies the discrepancy between a model's performance on observed examples and its performance on unseen instances. 


\begin{definition}[Generalization Gap]
\label{def_gen_gap}
Let $(X, Y) \sim \mathcal{D}$ and let $T = T(X)$ be an encoder output independent of the training data. Given a loss $\ell: \mathcal{Y}\times\mathcal{Y}\to\mathbb{R}_{\ge0}$ and predictor $f:\mathcal{T}\to\mathcal{Y}$, the \emph{generalization gap} is:
\[
\Delta \;=\; \mathbb{E}_{(X,Y)\sim\mathcal{D}}\bigl[\ell(f(T),Y)\bigr]
- \frac{1}{n}\sum_{i=1}^n \ell\bigl(f(t_i),y_i\bigr),
\]
where $(x_i,y_i)$ are i.i.d.\ samples from $\mathcal{D}$ and $t_i=T(x_i)$.
\end{definition}

This is particularly relevant to our problem, where a demonstration is provided and the agent is expected to solve a different, unseen task. Inspired by existing literature in deep learning theory \cite{xu2017information}, we have Theorem~\ref{xu_bound}. Further details are provided in the Appendix.

\begin{theorem}[]
\label{xu_bound}
Let the encoder $p(t\mid x)$ be fixed independently of the training data. Then the generalization gap is bounded by:
\[
\Delta \;\le\;\widetilde{O}\!\Bigl(\sqrt{\tfrac{I(X;T)+1}{n}}\Bigr),
\]
where $I(X;T)$ is the mutual information between $X$ and $T$, and $n$ is the number of training samples.
\end{theorem}

Theorem~\ref{xu_bound} provides a bound on the generalization gap, directly illustrating the benefit of compressing demonstrations to their transferable knowledge. In our setting, let an encoder compress a demonstration \( D \) into its transferable knowledge representation \( \mathit{TK} \). Then, the generalization gap is bounded by:
\[
\Delta(\mathit{TK}) \le \widetilde{O}\!\left(\sqrt{\tfrac{I(D;\mathit{TK}) + 1}{n}}\right),
\]
In contrast, if no such compression is applied, i.e., we set \( \mathit{TK} = D \), then \( I(D;\mathit{TK}) = I(D;D) = H(D) \), yielding the bound:
\[
\Delta(D) \le \widetilde{O}\!\left(\sqrt{\tfrac{H(D) + 1}{n}}\right).
\]

Since any non-trivial encoder will discard some non-essential information, we have \( I(D;\mathit{TK}) \le H(D) \), which implies \( \Delta(\mathit{TK}) \le \Delta(D) \). That means, as soon as \( \mathit{TK} \) omits even a small amount of information from \( D \), the generalization gap bound becomes strictly tighter. In other words, compressing demonstrations to their transferable components yields a provably stronger high-probability guarantee on the model’s performance.


Furthermore, we show that selecting demonstrations according to our proposed criterion (Equation~\ref{eq:criterion}) leads to a tighter generalization bound. To support this claim, we build our analysis on Theorem~\ref{kawaguchi_bound}, drawing on the theoretical framework from \cite{kawaguchi2023does}; additional details are provided in the Appendix.

\begin{theorem}[]
\label{kawaguchi_bound}
Let \(\varphi\) be a fixed encoder mapping each input \(X\) to a representation \(T = \varphi(X)\), and let \(\Delta\) denote the generalization gap of a classifier built on top of \(Z\).  Then, for a dataset of size \(n\), the generalization gap is bounded by:
\[
  \Delta \;\le\;\widetilde O\!\Bigl(\sqrt{\tfrac{I(X;T\mid Y)}{n}}\Bigr).
\]
\end{theorem}

In our setting, Notably, when \(\beta = 1\), the criterion in Equation~\ref{eq:criterion} can be reduced to:
\[
  \mathcal{J}_i = I(d_i;TK_i) - \,I(TK_i;A_t) = I(d_i; TK_i \mid A_t),
\]
with the details provided in the Appendix. Therefore, suppose we have two candidate demonstrations \(d_i\) and \(d_j\), and the selection criterion favors \(d_i\), i.e., \(I(d_i; TK_i \mid A_t) < I(d_j; TK_j \mid A_t)\). According to the generalization bound from Theorem~\ref{kawaguchi_bound}, which scales with \(I(d; TK \mid A_t)\), we obtain:
\[
\widetilde{O}\!\left(\sqrt{\tfrac{I(d_i; TK_i \mid A_t)}{n}}\right) < \widetilde{O}\!\left(\sqrt{\tfrac{I(d_j; TK_j \mid A_t)}{n}}\right).
\]

Thus, selecting demonstrations based on our proposed criterion provably yields a tighter generalization bound. This result reinforces our central claim: our method enhances in-context learning performance by isolating transferable knowledge and reducing the influence of task-specific noise.




\subsection{Transferable Knowledge Estimation and Demonstration Selection}

At each timestep \( t \), our goal is to select a demonstration \( d \) from a pool of candidates \( \mathcal{D} \) using the criterion in Equation~\ref{eq:criterion}. Since the mutual information terms are generally intractable to compute directly, we introduce practical strategies to approximate them.

First, we employ a pre-trained LLM as the \emph{Knowledge Retriever}, inducing a stochastic channel \( d \mapsto TK_d \) with fixed capacity. This leads to a constant mutual information \( I(d; TK_d) \) across all \( d \in \mathcal{D} \), simplifying the objective to maximizing \( I\bigl(TK_d; A_t\bigr) \), which encourages selecting demonstrations whose transferable knowledge is most predictive of the next action.

Then, since \( A_t \) is the action to be predicted, it is not observable at the time of demonstration selection. To address this, we employ the same \emph{Knowledge Retriever} to extract the anticipated transferable knowledge from the current agent context \( H_t \) denoted as \( TK_t \), as a proxy of \( A_t \). Then, to estimate the mutual information \( I(TK_d, TK_t) \), we apply the InfoNCE lower bound \cite{oord2018representation}, which leads to the following retrieval objective:
\[
d^\star \approx \arg\max_{d \in \mathcal{D}} \; \log\frac{\exp\!\bigl(\mathrm{sim}(TK_d,TK_t)\bigr)}%
{\sum_{d'\in\mathcal{D}}\exp\!\bigl(\mathrm{sim}(TK_{d'},TK_t)\bigr)},
\]
where \(\mathrm{sim}(\cdot,\cdot)\) denotes a similarity function, computed using cosine similarity. These approximations effectively bridge the gap between our demo selection criteria formulation (Equation~\ref{eq:criterion}) and a practical, efficient implementation for dynamic demonstration selection.

\section{Experiment}

\subsection{Experimental Setting}
\label{exp_setting}

\textbf{Tasks} Following prior work \cite{yao2023react,shinn2023reflexion,zhou2023language}, we evaluate our method on two diverse and challenging domains, including reasoning tasks and sequential decision-making tasks.
For reasoning tasks, we evaluate on HotpotQA\cite{yang2018hotpotqa}, a multi-hop question answering dataset, whereas for sequential decision making tasks, we evaluate on Webshop~\cite{yao2022webshop}, an interactive web-based shopping environment, and AlfWorld \cite{shridhar2020alfworld}, a text-based embodied decision-making environment. Detailed descriptions of the settings for each task are elaborated in Sections~\ref{exp_hotpotqa}–\ref{exp_alfworld}.

\textbf{Baselines}
We integrate our proposed method with several existing agentic frameworks, including ReAct \cite{yao2023react}, Reflexion \cite{shinn2023reflexion}, and LATS \cite{zhou2023language}. For each baseline, we adhere strictly to the original implementation details and hyperparameter settings to ensure fair and controlled comparisons. Specifically, for Reflexion, we use a trial count of 5, and for LATS, we adopt the LATS(ReAct) configuration as described in the original paper.

\textbf{LLM Agent}
In line with previous work \cite{shinn2023reflexion}, all experiments are conducted using the \textit{gpt-3.5-turbo} model. To ensure fair comparison, we reproduce some baseline results using \textit{gpt-3.5-turbo} when the original work used models such as \textit{text-davinci-002}, which are no longer supported. The reproduced results are consistent in magnitude with the original reports.

\textbf{Knowledge Retriever}
To extract transferable knowledge, we employ the open-source language model \textit{gemma-2-2b-it} \cite{team2024gemma} as the pre-trained encoder. Full details of the prompting strategy used to derive \text{TK} are provided in the Appendix.

\textbf{Demonstration Pool Construction}
Our method requires a pool of candidate demonstrations from which to retrieve. For each task, we construct this pool by running a small subset of task instances using the respective baseline agent and collecting only the successful trajectories. During evaluation, we use the remaining, non-overlapping task instances to assess performance.

\begin{wraptable}{R}{5cm}
  \vspace{-10pt}
  \centering
  \caption{Performance on HotpotQA. The number denotes the percentage of Exact Match (EM).}
  \small
  \begin{tabular}{l | l }
    \toprule
    & EM $\uparrow$ \\
    
    \midrule
    
    ReAct & 32.1 \\
    ReAct + DICE & 41.4 \textcolor{teal}{$\uparrow$(+9.3)}\\
    
    \midrule

    Reflexion & 51.6 \\
    Reflexion + DICE & 58.9 \textcolor{teal}{$\uparrow$(+7.3)}\\
    
    \midrule

    LATS & 63.3 \\
    LATS + DICE & 71.4 \textcolor{teal}{$\uparrow$(+8.1)}\\
    
    \bottomrule
  \end{tabular}
  \label{tab:result_hotpotqa}
  \vspace{-5pt}
\end{wraptable}

\subsection{Reasoning: HotpotQA}
\label{exp_hotpotqa}
\textbf{Setup} HotPotQA \cite{yang2018hotpotqa} is a multi-hop question answering benchmark based on Wikipedia, designed to test an agent's ability to retrieve and reason over multiple documents to answer complex questions. Following the setup from \cite{yao2023react}, we use a simplified Wikipedia environment that supports three interaction primitives: Search[entity], Lookup[string] and Finish[answer]. Performance is measured using the standard Exact Match (EM) score. We evaluate all methods on a 500-question subset. For other settings, such as the number of few-shot demonstrations, we adhere strictly to the exact settings specified in each baseline method.

\textbf{Result} Our experimental results are presented in Table~\ref{tab:result_hotpotqa}. Across all baselines, integrating our method yields consistent and non-trivial improvements in Exact Match (EM) score. These gains highlight the effectiveness of our approach in enhancing agentic reasoning across different frameworks, demonstrating its generality and robustness. Notably, the improvements hold across agent architectures, suggesting our method offers benefits orthogonal to model-specific optimizations.

\subsection{Sequential decision making: ALFWorld \& Webshop}
\label{exp_alfworld}

\begin{wraptable}{R}{8cm}
\vspace{-10pt}
  \centering
  \caption{Score and success rate (SR) on Webshop. Our method consistently improves performance across all baseline methods by a significant margin.}
  \small
  \begin{tabular}{l | l l}
    \toprule
    & Score $\uparrow$ & SR $\uparrow$ \\
    
    \midrule
    
    ReAct & 53.8 & 28.0 \\
    ReAct + DICE & 58.2 \textcolor{teal}{$\uparrow$(+4.4)} & 35.0 \textcolor{teal}{$\uparrow$(+7.0)} \\

    \midrule
    
    Reflexion & 64.2 & 35.0 \\
    Reflexion + DICE & 67.4 \textcolor{teal}{$\uparrow$(+3.2)} & 38.2 \textcolor{teal}{$\uparrow$(+3.2)} \\
    
    \midrule

    LATS & 75.9 & 38.0 \\
    LATS + DICE & 77.1 \textcolor{teal}{$\uparrow$(+1.2)} & 39.5 \textcolor{teal}{$\uparrow$(+1.5)} \\ 
    
    \bottomrule
  \end{tabular}
  \label{tab:result_webshop}
  \vspace{-5pt}
\end{wraptable}

\textbf{AlfWorld} \cite{shridhar2020alfworld} is a suite of interactive, text-based environments adapted from the ALFRED \cite{shridhar2020alfred} benchmark, where an agent completes multi-step household tasks via natural language commands. These tasks span six categories: Pick, Clean, Heat, Cool, Look, and Pick Two. The agent interacts with a simulated home by issuing actions like \texttt{go to}, \texttt{take}, or \texttt{use}, navigating complex environments with multiple rooms and objects. Each task may require over 50 steps, demanding effective planning, subgoal tracking, and commonsense reasoning. Following prior work, we evaluate our method on 134 unseen evaluation task instances using Success Rate (SR) as the performance metric.

\textbf{WebShop} \cite{yao2022webshop} is a language-based environment simulating an online shopping platform, where agents are tasked with selecting a product that satisfies a user’s natural language query, such as ``a black desk chair under \$100 with lumbar support''. Agents interact via search queries, button selections, and page navigation, simulating web browsing behavior. Performance is measured using two metrics: \textbf{success rate}, and the \textbf{average score}, which captures the average reward obtained across episodes. In line with previous work\cite{yao2023react,shinn2023reflexion}, we use the test set of 500 shopping tasks for evaluation and maintain consistency with established action space definitions and prompting strategies.

\textbf{Result} The results are shown in Table~\ref{tab:result_alfworld} and Table~\ref{tab:result_webshop} for AlfWorld and Webshop respectively. The result shows that integrating our method into existing agentic frameworks leads to consistent performance improvements across both benchmarks. On AlfWorld, we observe substantial gains in success rates across all subcategories, especially in more challenging tasks like “Clean” and “Pick 2”. Similarly, on Webshop, our approach improves both average score and success rate across all baselines. These results demonstrate the broad applicability and effectiveness of our method in enhancing agent performance in diverse and complex environments.

\begin{table*}[t]
  \centering
  \caption{Results on AlfWorld with Success Rates (SR) across subcategories. Our method consistently improves performance across all subcategories and achieves a significant margin over the baseline.}
  \small
  \begin{tabular}{l | l l l l l l | l }
    \toprule
    & Pick & Clean & Heat & Cool & Look & Pick 2 & All \\
    
    \midrule
    
    ReAct & 66.7 & 38.7 & 82.6 & 76.2 & 55.6 & 23.5 & 57.5 \\
    ReAct + DICE & 79.2 & 51.6 & 87.0 & 76.2 & 66.7 & 47.1 & 67.9 \textcolor{teal}{$\uparrow$(+10.4)} \\
    
    \midrule

    Reflexion & 83.3 & 58.1 & 91.3 & 90.5 & 72.2 & 52.9 & 74.6 \\
    Reflexion + DICE & 87.5 & 71.0 & 95.7 & 90.5 & 83.3 & 64.7 & 82.1 \textcolor{teal}{$\uparrow$(+7.5)}\\
    
    \bottomrule
  \end{tabular}
  \label{tab:result_alfworld}
\end{table*}

\subsection{Ablation Study \& Other Comparisons}

\begin{wraptable}{R}{8cm}
  \vspace{-10pt}
  \centering
  \caption{Comparison on Taskwise \& Stepwise demo selection methods, evaluated by  Exact Match(EM) on HotpotQA, SR for AlfWorld and Webshop.}
  \small
  \begin{tabular}{l | c c c}
    \toprule
    & HotpotQA & AlfWorld & Webshop \\
    
    \midrule
    
    ReAct & 32.1 & 57.5 & 28.0 \\

    \midrule

    KATE & 34.7 & 58.2 & 30.5 \\
    EPR & 36.5 & 60.3 & 30.1 \\

    \midrule
    
    DICE (Taskwise) & 36.3 & 61.2 & 31.3 \\
    DICE (Stepwise) & \textbf{41.4} & \textbf{67.9} & \textbf{35.0} \\
    
    \bottomrule
  \end{tabular}
  \label{tab:result_task_obs_lvl}
\end{wraptable}

A key feature of our method is stepwise demonstration selection, i.e., dynamically choosing different demonstrations at each timestep as the agent progresses through a task. To evaluate the benefit of this design, we conduct an ablation study comparing it with a task-level variant, where a fixed set of demonstrations selected by DICE at timestep 0 is used throughout the task.

We also compare against established selection methods including KATE \cite{liu2021makes}, which uses a $k$NN-based strategy to select relevant demonstrations, and EPR \cite{rubin2021learning}, which trains an additional module to facilitate selection. While both methods were originally proposed for one-step question answering, they lack adaptation for multi-step agentic settings. Thus, we treat them as task-level selectors in our experiments.

The results, summarized in Table~\ref{tab:result_task_obs_lvl}, highlight the following: (1) Taskwise DICE outperforms existing methods, underscoring the importance of isolating transferable knowledge (TK). While EPR performs comparably, it requires additional training, whereas our method is entirely training-free.(2) Stepwise DICE consistently outperforms its taskwise variant, demonstrating the value of adapting demonstrations at each reasoning step. This confirms the effectiveness of our approach and shows that stepwise selection yields meaningful gains beyond what static, task-level selection can achieve.

\subsection{Further Analysis of DICE Performance}
\label{exp_analysis}

Beyond the main evaluations stated in previous sections, we conduct additional analyses to better understand the effectiveness and characteristics of our method. The results in this section are from on HotpotQA, evaluated by Exact Match (EM), results on other tasks are provided in the Appendix.

\begin{figure}[t]
    \centering
    \begin{subfigure}[t]{0.31\textwidth}
        \centering
        \includegraphics[width=\linewidth]{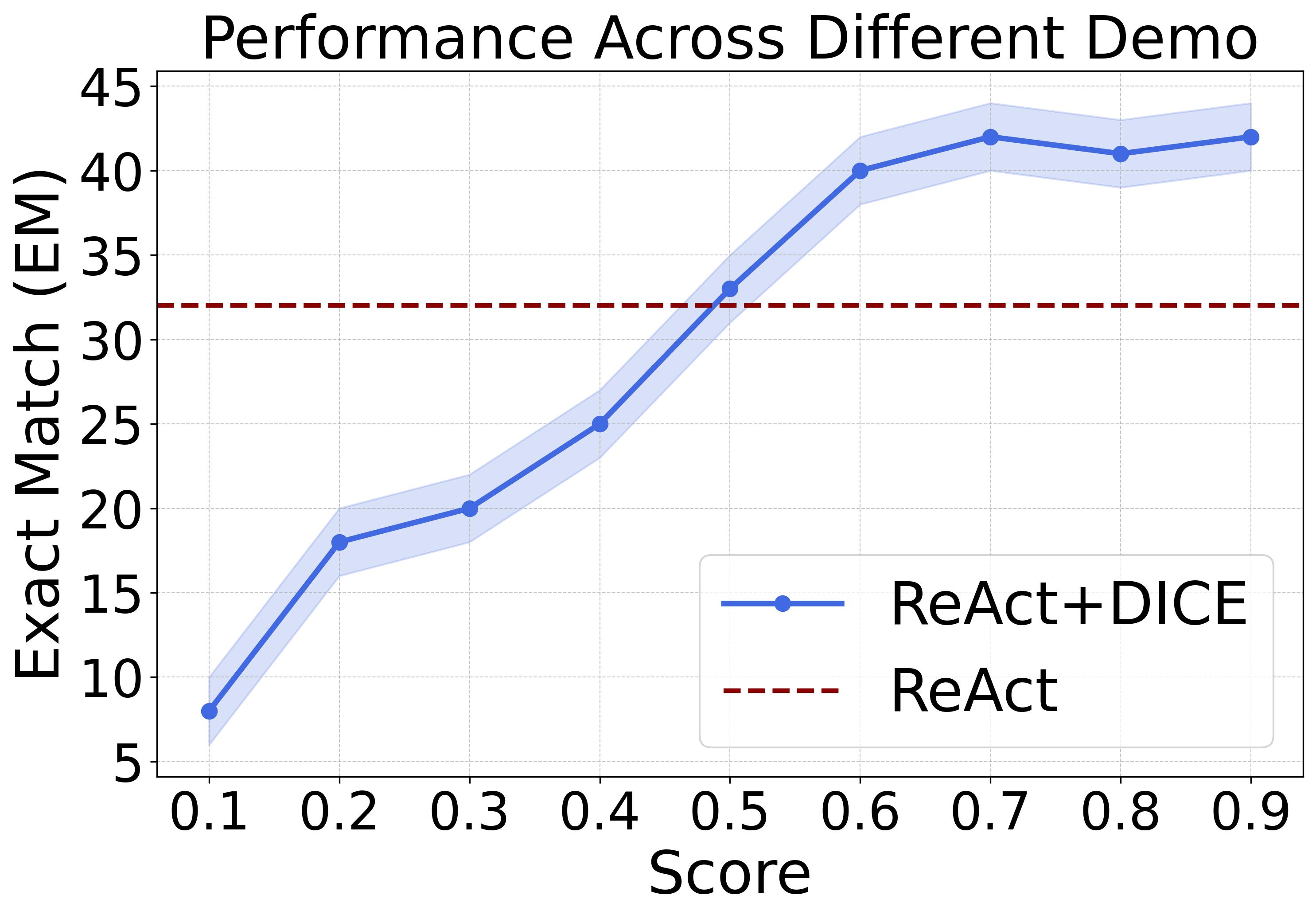}
        \caption{}
        \label{fig:performance_vs_score}
    \end{subfigure}
    \hfill
    \begin{subfigure}[t]{0.32\textwidth}
        \centering
        \includegraphics[width=\linewidth]{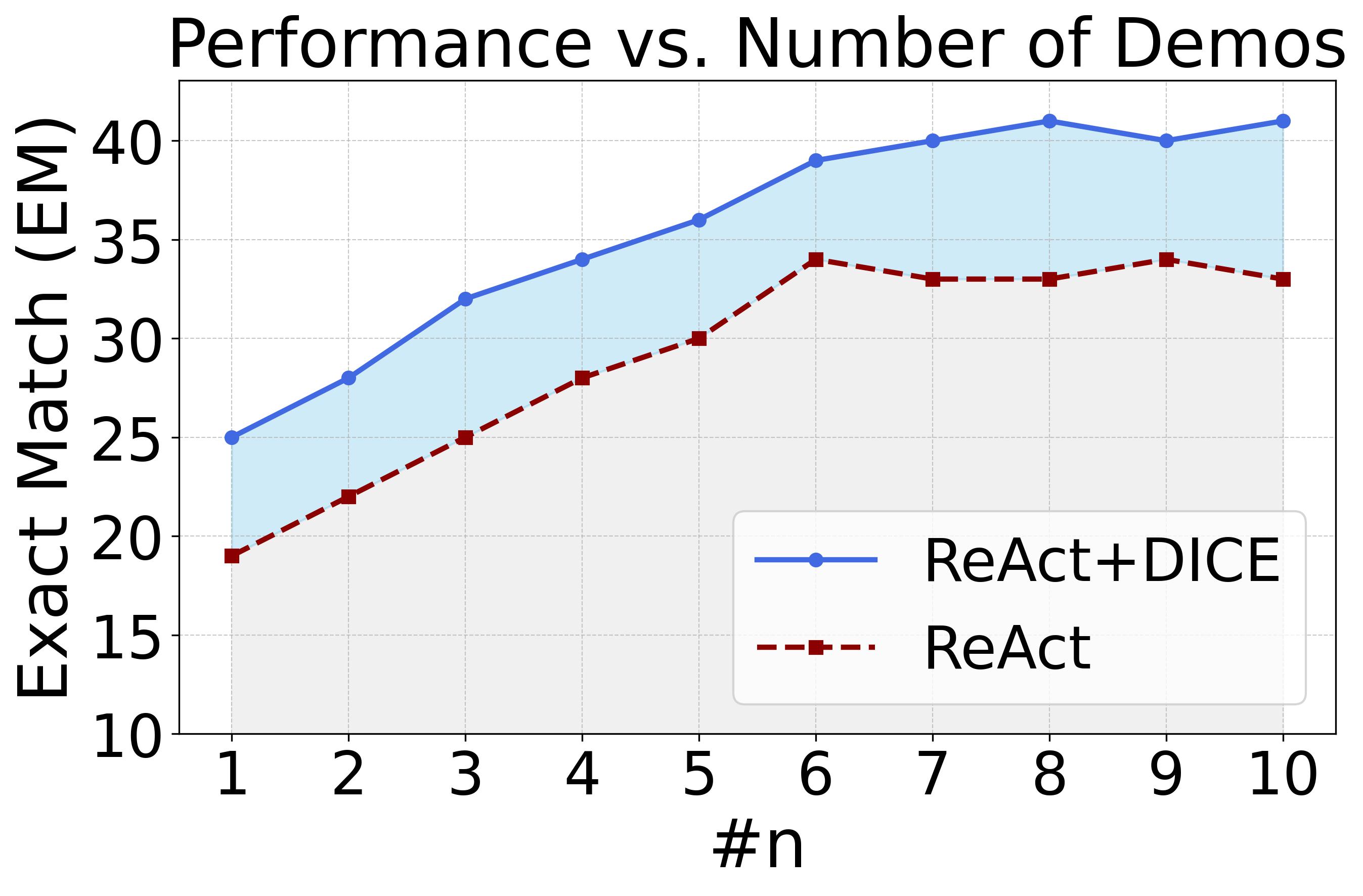}
        \caption{}
        \label{fig:performance_vs_num_demo}
    \end{subfigure}
    \hfill
    \begin{subfigure}[t]{0.32\textwidth}
        \centering
        \includegraphics[width=\linewidth]{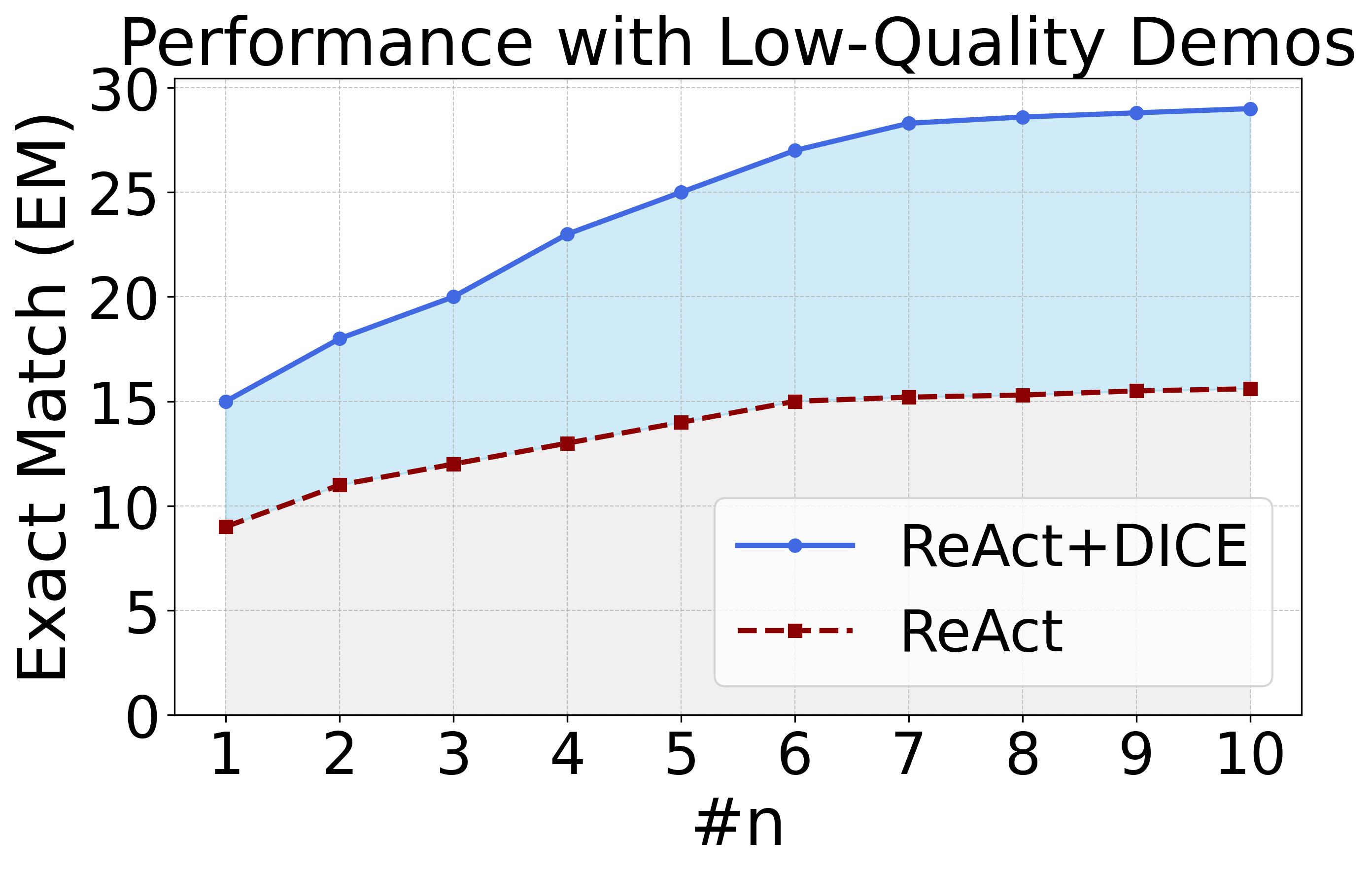}
        \caption{}
        \label{fig:performance_bad_demo}
    \end{subfigure}
    \caption{(a) Using demonstrations with higher average score in DICE results in better performance. (b) DICE outperforms standard ICL consistantly with the same number of demonstrations, and matches its performance with significantly fewer examples. (c) DICE maintains an even stronger margin over standard ICL with only low-quality demos, demonstrating robustness and the ability to extract transferable knowledge from suboptimal inputs.}
    \label{fig:analysis_figure}
\end{figure}

\paragraph{How does better in-context example benefit improve performance?}
Our method assigns a relevance score between 0 and 1 to each demo-target pair, reflecting how well a demonstration aligns with the target under our proposed selection criterion. To evaluate whether these scores correspond to actual utility, we group episodes based on the average score of their selected demonstrations and plot performance as a function of demonstration quality (Figure~\ref{fig:performance_vs_score}). A horizontal red dashed line indicates the baseline performance under random demo selection (i.e., ReAct). As the figure shows, selecting higher-scoring demonstrations consistently leads to improved performance, clearly illustrating that our scoring metric is predictive of ICL effectiveness.

\paragraph{How does our method reduce the number of required demonstrations?}
By selecting higher-quality demonstrations, our method achieves comparable or better performance using fewer examples. As shown in Figure~\ref{fig:performance_vs_num_demo}, DICE consistently outperforms standard ICL across different numbers of demonstrations. Notably, with just 3 selected examples, DICE matches the performance of standard ICL that uses 6 randomly chosen ones. This highlights the efficiency of our approach in reducing reliance on large numbers of demonstrations without sacrificing effectiveness.

\paragraph{What if there are no relevant demonstrations available?}
In scenarios where no highly relevant demonstrations are available, it is important to assess whether our method still provides benefits. To simulate this, we constrain the demonstration pool to only include examples with a relevance score below 0.5 and compare the performance of DICE and standard ICL in Figure~\ref{fig:performance_bad_demo}. Interestingly, we observe that the performance gap in favor of DICE becomes even larger under these challenging conditions. This result suggests that our method remains effective even with suboptimal demonstrations—likely due to its principled focus on maximizing transferable knowledge, which enables better utilization of available examples despite their lower overall quality.

\subsection{Case Study}
\label{case_study}
\begin{figure*}[t]
  \centering
  \includegraphics[width=0.9\linewidth]{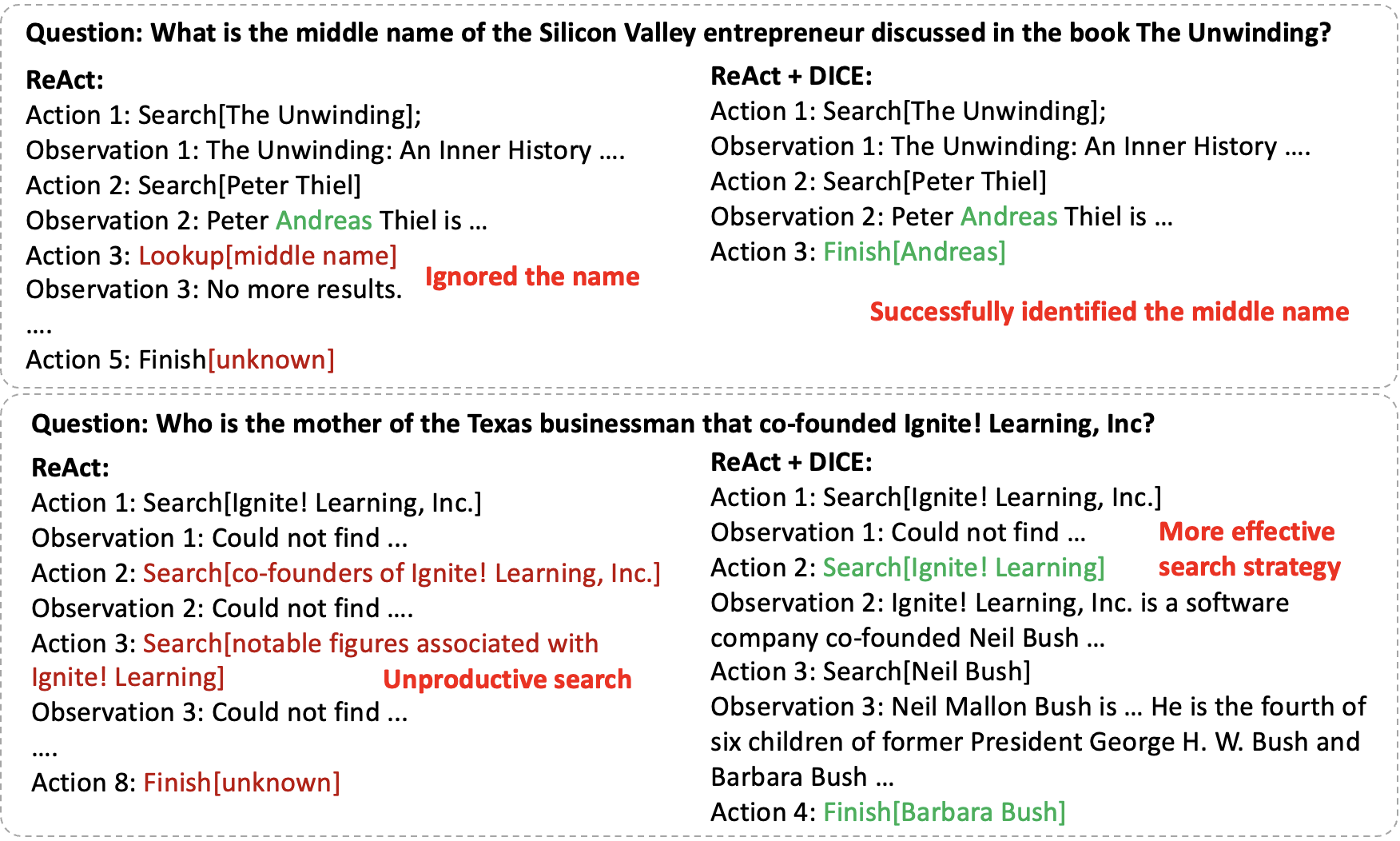}
  \caption{Case studies. \textbf{Top}: Without DICE, the agent overlooks the correct answer already present in Observation 2 and unnecessarily attempts a \texttt{Lookup[middle name]}, resulting in failure. With DICE, the agent correctly extracts the middle name directly from the observation. \textbf{Bottom}: Without DICE, the agent issues a series of unproductive search queries, while DICE enables the agent to identify and search for the correct term by the second step, leading to a successful answer.}
  \label{fig:case_study}
\end{figure*}

To qualitatively illustrate the advantages of our approach, we present some representative examples from the experiments, as shown in Figure~\ref{fig:case_study}. In the first case, ReAct fails to identify the answer even though it appears directly in the observation, while ReAct + DICE correctly extracts the answer without unnecessary steps. This suggests that our method is better at leveraging useful information already found during the problem-solving process. In the second case, ReAct issues several overly specific and unproductive search queries, while ReAct + DICE quickly adjusts its strategy to a more effective search term and successfully retrieves the correct answer. This demonstrates our method's ability to guide the agent toward more relevant queries when the initial attempt fails. These examples highlight how integrating DICE into the agent framework leads to more accurate and efficient reasoning in challenging scenarios, and improves reasoning and decision-making.

\section{Related Work}

\textbf{Agentic Framework} 
Agentic frameworks empower LLMs to solve long-horizon tasks by enabling observation reading, natural-language planning, and tool use \citep{yao2023react, schick2023toolformer, shinn2023reflexion, zhou2023language, ma2023laser, zhao2024expel, nguyen2024gui}. ReAct \citep{yao2023react} shows that interleaving reasoning and actions improves performance over purely reasoning- or action-based approaches. Toolformer \citep{schick2023toolformer} enables zero-shot API usage by letting the model decide when to invoke tools without extra supervision. Reflexion \citep{shinn2023reflexion} introduces a self-critique loop to reduce repeated errors. LATS \citep{zhou2023language} unifies reasoning, acting, and planning via a tree search algorithm.

\textbf{In-Context Learning \& Demo Selection}
In-context learning (ICL) enables language models to solve new tasks by conditioning on a few input–output examples without weight updates \citep{brown2020language}. Recent works enhance ICL via specialized pretraining \citep{gu2023pre,li2024mend,shi2023context} or intermediate training stages \citep{min2021metaicl,iyer2022opt,wang2022super,wei2021finetuned,chung2024scaling}. However, the effectiveness of ICL is highly sensitive to the choice and order of demonstrations, leading to research on calibration \citep{min2022rethinking, zhao2021calibrate}, ordering \citep{lu2022fantastically}, and rationale-based prompting \citep{wei2022cot}. Adaptive methods address this by selecting examples per query using compression objectives \citep{wu2023self} or feedback-trained retrievers \citep{wang2024learning}. 
Theoretical studies also analyze the effectiveness of ICL \citep{akyurek2023iclgrad, dai2023can, li2024closeness}. 

Recent works have explored various demo selection strategies \citep{dong2022survey}. Unsupervised methods typically retrieve nearest neighbours based on input similarity \citep{liu2021makes, tanwar2023multilingual, qin2023context}, or use model output scores as selection criteria \citep{nguyen2023context, li2023finding, wu2022self}. Mutual information \citep{sorensen2022information}, perplexity \citep{gonen2022demystifying} and graphs \cite{su2022selective} have also proven effective for prompt selection.
Alternatively, some approaches train auxiliary models to guide selection \citep{rubin2021learning, ye2023compositional, wang2023large, van2024context,zhang2022active,wang2024demonstration}. However, these require supervision and additional training, limiting their applicability in practice. Other approaches refine prompts iteratively or rank examples by difficulty or uncertainty \citep{qin2024context, vu2024curriculum, xu2024misconfidence}, and \citep{zhao2023dynamic} dynamically adjusts prompt length to avoid performance drops from excessive examples.
Other techniques such as Demonstration Reformatting \citep{kim2022self,hao2022structured,liu2023context} and  Demonstration Ordering \citep{liu2024let,lu2021fantastically} are also explored.
However, these works focus on non-agentic settings. While recent efforts target agentic tasks \citep{lutz2024wilbur}, they are domain-specific and lack theoretical grounding. In contrast, our DICE framework offers a general, theoretically grounded solution for dynamic demonstration selection in multi-step agentic tasks.


\section{Conclusion}

We introduced DICE, a dynamic and theoretically grounded in-context learning framework designed to enhance the performance and robustness of LLM-based agents in multi-step reasoning and tool-use tasks. By modeling the demonstration selection process through a causal lens, we identified how non-transferable knowledge in examples can introduce spurious dependencies and hurt generalization. Our proposed method dynamically selects the most relevant demonstrations at each reasoning step, maximizing transferable knowledge while operating entirely at inference time and requiring no additional training. Empirical results across diverse benchmarks and agentic frameworks validate the effectiveness of our approach, showing consistent gains over strong baselines. One limitation of our work is that we only explore a single instantiation of the framework; more expressive implementations—such as incorporating a trainable encoder to capture latent transferable knowledge—are left for future exploration. A broader impact of our approach is the potential to enable user-customized agents that learn from tailored examples, supporting more intuitive and adaptive human-AI interaction.


\bibliographystyle{plainnat}
\bibliography{ref}

\end{document}